%% file: main.tex
\documentclass[conference]{IEEEtran}
\usepackage[shortlabels]{enumitem}
\usepackage[nospace,noadjust]{cite}
\usepackage{float}
\usepackage{graphicx} 
\usepackage{epsfig} 
\usepackage{amsmath,amsthm,amssymb}  
\usepackage{float}
\usepackage{cases,setspace,adjustbox}
\usepackage{array}
\usepackage[update,prepend]{epstopdf}
\usepackage{caption}
\usepackage{eqparbox}
\usepackage{url}
\usepackage{caption, subcaption}
\usepackage{xcolor}
\usepackage{mathtools}
\usepackage{bbm}
\usepackage{multirow}
\usepackage{multicol}

\usepackage[acronym]{glossaries}
\usepackage{comment}
\usepackage{xcolor}
\newcommand*\circled[1]{\kern-2.5em%
  \put(0,4){\color{white}\circle*{11}}\put(0,4){\circle{11}}%
  \put(-3,0){\color{black}\large#1}~~}
\input{glossary}
\begin{document}
\title{O-RAN-Native Jamming Detection Under Network Heterogeneity}
\author{Sneihil Gopal$^{\ddagger\dagger}$, Fernando J. Cintr\'{o}n$^{\star}$, and Richard A. Rouil$^{\star}$\\
$^{\ddagger}$PREP Associate, National Institute of Standards and Technology (NIST), USA\\
$^{\dagger}$Department of Physics, Georgetown University, USA\\
$^{\star}$National Institute of Standards and Technology (NIST), USA.\\
Emails: \{sneihil.gopal, fernando.cintron, richard.rouil\}@nist.gov}
\maketitle
\input{abstract}
\input{Introduction_V2}
\input{Related_Work_v1}
\input{Methodology_V1}
\input{Experimental_Setup_V1}
\input{Results_V1}
\input{System_Integration_V1}
\input{Conclusion_V1}
\input{Acknowledgement}
\begin{spacing}{}
\bibliographystyle{IEEEtran}
\bibliography{references}
\end{spacing}
\end{document}

%% file: glossary.tex
\setacronymstyle{long-short}

\newacronym{5G}{5G}{fifth-generation}
\newacronym{6G}{6G}{sixth-generation}
\newacronym{c2}{C2}{command-and-control}
\newacronym{1dcnn}{1D-CNN}{one-dimensional Convolutional Neural Network}
\newacronym{AR}{AR}{Augmented Reality}
\newacronym{ai}{AI}{Artificial Intelligence}
\newacronym{arima}{ARIMA}{Autoregressive Integrated Moving Average}

\newacronym{cdf}{CDF}{cumulative distribution function}
\newacronym{cnn}{CNN}{Convolutional Neural Network}
\newacronym{comp}{CoMP}{Coordinated Multi-Point}
\newacronym{convlstm}{ConvLSTM}{Convolutional LSTM}
\newacronym{cp}{CP}{Cyclic Prefix}
\newacronym{cpd}{CPD}{Change Point Detection}
\newacronym{crs}{CRS}{Cell Specific
Reference Signal}
\newacronym{ctl}{CTL}{Communications Technology Laboratory}
\newacronym{CU}{CU}{Centralized Unit}
\newacronym{cusum}{CUSUM}{Cumulative Sum Control Chart}

\newacronym{d2d}{D2D}{Device-to-Device}
\newacronym{dc}{DC}{dual connectivity}
\newacronym{dci}{DCI}{Downlink Control Information}
\newacronym{DL}{DL}{deep learning}
\newacronym{dl}{DL}{downlink}
\newacronym{dsrc}{DSRC}{Dedicated short-range communications}
\newacronym{dss}{DSS}{Dynamic Spectrum Sharing}
\newacronym{DU}{DU}{Distributed Unit}

\newacronym{ets}{ETS}{Exponential Smoothing}
\newacronym{embb}{eMBB}{enhanced mobile broadband}
\newacronym{enb}{eNB}{Evolved Node-B}
\newacronym{epc}{EPC}{Evolved Packet Core}
\newacronym{emtc}{eMTC}{enhanced MTC}

\newacronym{fdd}{FDD}{Frequency Domain Duplex}
\newacronym{fdm}{FDM}{Frequency Domain Multiplexing}
\newacronym{firstnet}{FirstNet}{First Responder Network Authority}

\newacronym{gan}{GAN}{Generative Adversarial Network}
\newacronym{gps}{GPS}{Global Positioning System}
\newacronym{gnb}{gNB}{Next Generation Node-B}
\newacronym{gbps}{Gb/s}{gigabits per second}
\newacronym{gru}{GRU}{Gated Recurrent Unit}

\newacronym{harq}{HARQ}{Hybrid Automatic Repeat reQuest}
\newacronym{haps}{HAPS}{High-Altitude Platform Stations}

\newacronym{icmp}{ICMP}{Internet Control Message Protocol}
\newacronym{iot}{IoT}{Internet of Things}
\newacronym{iiot}{IIoT}{Industrial Internet of Things}
\newacronym{iomt}{IoMT}{Internet of Medical Things}
\newacronym{irs}{IRS}{Intelligent Reflecting Surfaces}
\newacronym{its}{ITS}{Intelligent Transport Systems}
\newacronym{itu}{ITU}{International Telecom Union}

\newacronym{kpi}{KPI}{Key Performance Indicator}
\newacronym{kpis}{KPIs}{Key Performance Indicators}

\newacronym{loo}{LOO}{leave-one-out}
\newacronym{lte}{LTE}{Long Term Evolution}
\newacronym{lteapro}{LTE-A~Pro}{LTE Advanced Pro}
\newacronym{lstm}{LSTM}{Long Short-Term Memory}
\newacronym{lr}{LR}{Logistic Regression}

\newacronym{mad}{MAD}{Median Absolute Deviation}
\newacronym{mcs}{MCS}{Modulation and Coding Scheme}
\newacronym{mcptt}{MCPTT}{Mission-Critical Push-to-Talk}
\newacronym{mmimo}{mMIMO}{massive Multiple-Input Multiple-Output}
\newacronym{mmtc}{mMTC}{massive machine-type communication}
\newacronym{mtc}{MTC}{Machine-Type Communication}
\newacronym{mlp}{MLP}{Multilayer Perceptron}
\newacronym{ma}{MA}{Moving Average}
\newacronym{mm}{MM}{Moving Median}
\newacronym{ml}{ML}{machine learning}

\newacronym{nist}{NIST}{National Institute of Standards and Technology}
\newacronym{nbiot}{NB-IoT}{Narrowband-Internet of Things}
\newacronym{nr}{NR}{New Radio}
\newacronym{ntia}{NTIA}{National Telecommunications and Information Administration}
\newacronym{ntn}{NTN}{Non-Terrestrial Networks}

\newacronym{oran}{O-RAN}{Open Radio Access Network}
\newacronym{owl}{OWL}{Online Watcher of LTE}

\newacronym{pelt}{PELT}{Pruned Exact Linear Time}
\newacronym{prosas}{ProSAS}{Proactive Spectrum Adaptation Scheme}
\newacronym{prb}{PRBs}{Physical Resource Blocks}
\newacronym{psc}{PSC}{Public Safety Communications}
\newacronym{pscch}{PSCCH}{Physical Sidelink Control Channel}
\newacronym{psfch}{PSFCH}{Physical Sidelink Feedback Channel}
\newacronym{pssch}{PSSCH}{Physical Sidelink Shared Channel}

\newacronym{qos}{QoS}{Quality of Service}

\newacronym{ran}{RAN}{Radio Access Network}
\newacronym{rApp}{rApp}{RAN Application}
\newacronym{rans}{RANs}{Radio Access Networks}
\newacronym{rat}{RAT}{Radio Access Technology}
\newacronym{re}{RE}{Resource Element}
\newacronym{rf}{RF}{Random Forest}
\newacronym{ric}{RIC}{RAN Intelligent Controller}
\newacronym{ris}{RIS}{Reconfigurable Intelligent Surfaces} 
\newacronym{relu}{ReLU}{Rectified Linear Unit}
\newacronym{rl}{RL}{Reinforcement Learning}
\newacronym{rmse}{RMSE}{root-mean-square error}
\newacronym{rnn}{RNN}{Recurrent Neural Network}
\newacronym{rssi}{RSSI}{Received Signal Strength Indicator}
\newacronym{rsrp}{RSRP}{Reference Signal Received Power}
\newacronym{RU}{RU}{Radio Unit}
\newacronym{rt}{RT}{real-time}

\newacronym{sdr}{SDR}{Software Defined Radio}
\newacronym{sem}{SEM}{Statistical Ensemble Method}
\newacronym{sl}{SL}{sidelink}
\newacronym{smo}{SMO}{Service and Management Orchestration}
\newacronym{snr}{SNR}{Signal-to-Noise Ratio}
\newacronym{ssb}{SSB}{Synchronization Signal Block}
\newacronym{svm}{SVM}{Support Vector Machine}

\newacronym{tcn}{TCN}{Temporal Convolutional Network}
\newacronym{td3}{TD3}{Twin Delayed Deep Deterministic Policy Gradient}
\newacronym{tdd}{TDD}{Time Domain Duplex}
\newacronym{tdm}{TDM}{Time Domain Multiplexing}
\newacronym{timegan}{TimeGAN}{Time-Series Generative Adversarial Network}
\newacronym{tr}{TR}{Technical Report}
\newacronym{thz}{THz}{terahertz}
\newacronym{TIP}{TIP}{Telecom Infra Project}

\newacronym{urllc}{URLLC}{ultra-reliable low-latency communication}
\newacronym{uav}{UAVs}{Unmanned Aerial Vehicles}
\newacronym{ue}{UE}{User Equipment}
\newacronym{ul}{UL}{uplink}

\newacronym{v2v}{V2V}{Vehicle-to-Vehicle}
\newacronym{v2x}{V2X}{Vehicle-to-Everything}
\newacronym{VR}{VR}{Vitual Reality}
\newacronym{vnf}{VNFs}{Virtualized Network Functions}
\newacronym{vm}{VMs}{Virtual Machines}

\newacronym{wnd}{WND}{Wireless Networks Division}
\newacronym{wlan}{WLAN}{Wireless Local Area Network}
\newacronym{xApp}{xApp}{eXtended application}
\newacronym{xr}{XR}{Extended Reality}
\newacronym{3gpp}{3GPP}{3rd Generation Partnership Project}

%% file: abstract.tex
\begin{abstract}
Radio-frequency jamming threatens cellular network operations in contested environments. The Open Radio Access Network (O-RAN) architecture exposes per-User Equipment (UE) telemetry through the E2 interface, enabling Key Performance Indicator (KPI)-based jamming detection without proprietary hardware. However, operational deployment demands detection methods that generalize across heterogeneous networks. We propose a Statistical Ensemble Method (SEM) combining adaptive Median Absolute Deviation (MAD)-based baseline detection, count-based anomaly detection, and abrupt change detection through a weighted voting system. SEM operates in two modes: a cold-start mode requiring no labeled data for immediate deployment, and a supervised mode where parameters are refined offline via grid search as labeled data accumulates. We evaluate SEM against five machine learning (ML) classifiers and five deep learning (DL) sequence models using approximately $\mathbf{25}$ hours of testbed data comprising logs from eight heterogeneous 5G environments, employing both single-environment and multi-environment leave-one-out (LOO) validation. Our analysis indicates that under conditions not encountered during training (distribution shift), learning-based models tend to lose recall and miss jamming events, whereas SEM, even in cold-start mode, maintains detection capabilities at the cost of false alarms. Furthermore, SEM's parameter optimization can be mapped to O-RAN's Non-RT RIC as an rApp for continuous refinement, allowing operators to transition to learning-based alternatives once corpus coverage is sufficient.
\end{abstract}

%% file: Introduction_V2.tex
\section{Introduction}
\label{sec:intro}
Radio-frequency jamming poses a growing threat to military and civilian cellular network operations. By overwhelming legitimate transmissions with broadband interference, a jammer can suppress reference signal measurements, degrade handover reliability, and cause complete service outages. These disruptions can have serious consequences for \gls{c2} communications and critical infrastructure in contested environments~\cite{pirayesh2022jamming}. The \gls{oran} architecture offers an effective platform for jamming detection by enabling the monitoring of per-\gls{ue} telemetry, including \gls{rsrp} statistics, \gls{snr}, and \gls{rssi}. This data is made available through standardized E2 interface reports, which can be accessed by detection applications in the \gls{ric} without the need for proprietary instrumentation~\cite{kryszkiewicz2023open,bogucka2025open}.

\Gls{ml} classifiers and \gls{DL} sequence models have shown strong performance in \gls{kpi}-based jamming detection during controlled evaluations~\cite{jere2023bayesian,hong2023hybrid,varotto2024detecting}. However, their effectiveness is constrained by two major challenges, particularly in military deployments. The first is \emph{distribution shift}: a model trained in one environment learns signal characteristics specific to that environment and may struggle when deployed under different operating conditions. The second is the \emph{cold-start problem}: learning-based models require labeled attack data before they can be trained for deployment~\cite{arjoune2020novel,gecgel2019jammer}. In a new deployment, such labeled data is typically unavailable, meaning that detection capabilities cannot be established until sufficient examples are collected and annotated.

To address these limitations, we propose a \gls{sem}. The core detectors of \gls{sem} are based on signal characteristics and employ well-defined default parameters, requiring no labeled data to begin operating. When labeled data becomes available, secondary parameters can be refined offline through grid search. This yields a two-mode design: a \emph{cold-start mode} for immediate detection and a \emph{supervised mode} for subsequent parameter refinement. We evaluate \gls{sem} against five \gls{ml} classifiers and five \gls{DL} sequence models using approximately $25$ hours of testbed data comprising eight heterogeneous \gls{5G} logs. The evaluation considers both single-environment and multi-environment \gls{loo} validation. The logs vary in \gls{ue} count, recording duration, jammer level, and burst pattern.

Across these heterogeneous conditions, the results reveal three main findings. First, the advantage of \gls{sem} is not fully reflected by its mean F1-score, which balances precision and recall. Instead, its strength lies in how it behaves under challenging conditions. Under distribution shift, learning-based models can lose recall and miss jamming events, whereas \gls{sem} maintains detection at the cost of additional false alarms. On the most challenging log, where jamming burst durations vary randomly, single-environment \gls{ml} and \gls{DL} achieve recalls of only $0.29$ and $0.47$, respectively, whereas \gls{sem} maintains a recall of $0.96$ even in cold-start mode. Second, multi-environment training substantially improves the performance of learning-based models compared with single-environment training. Specifically, the F1-score for \gls{ml} increases from $0.911$ to $0.962$, while that of \gls{DL} increases from $0.897$ to $0.956$. However, this improvement depends on the training corpus sufficiently representing the target deployment conditions, which cannot be guaranteed during initial rollout. Third, the results indicate that corpus coverage, rather than hyperparameter configuration alone, is a primary constraint on the generalization of learning-based methods across heterogeneous environments.

The dual-mode design of \gls{sem} aligns naturally with the \gls{oran} \gls{ric} architecture. 
In cold-start mode, SEM operates as an \gls{xApp} on the near-\gls{rt} \gls{ric}, receiving live \gls{kpi} telemetry through the E2 interface and allowing detection processing to begin immediately upon activation. As labeled incidents accumulate, the non-\gls{rt} \gls{ric} can host a parameter-optimization \gls{rApp} that refines \gls{sem}'s parameters through grid search. The optimized parameters can then be delivered through the A1 interface, allowing \gls{sem} to transition to supervised mode without interrupting live detection. Once the labeled dataset sufficiently represents the target deployment conditions, the \gls{rApp} can assess whether a learning-based alternative achieves performance comparable to \gls{sem} on validation data, allowing operators to determine whether a transition is appropriate. This deployment approach addresses the cold-start barrier that can hinder learning-based methods in new deployments.

The remainder of this paper is organized as follows. Section~\ref{sec:prior_work} reviews related work. Section~\ref{sec:methodology} describes \gls{sem} and the learning-based models. Section~\ref{sec:setup} presents the experimental setup. Section~\ref{sec:results} reports the results. Section~\ref{sec:integration} discusses \gls{oran} integration, and Section~\ref{sec:conclusion} concludes the paper.

%% file: Related_Work_v1.tex
\section{Related Work}
\label{sec:prior_work}

Jamming detection approaches span signal-level analysis, threshold-based methods, and \gls{kpi}-based statistical detection. Early work established that jamming produces characteristic degradation signatures (\gls{rsrp} drops, \gls{snr} variations, and \gls{rsrp} count loss) observable without physical-layer access~\cite{pirayesh2022jamming}. Threshold-based methods require manual calibration per site and do not adapt to changing conditions~\cite{szilagyi2012automatic}, while statistical ensemble and time-series approaches have shown promise for cellular anomaly detection~\cite{ciocarlie2013detecting,cheng2017time}. More recently, \gls{kpi}-based detection has been demonstrated in \gls{5G} \gls{nr} contexts~\cite{wang2023detecting,chiarello2021jamming} and within \gls{oran} architectures~\cite{kryszkiewicz2023open,bogucka2025open,hoffmann2023signaling}.

\gls{ml} and \gls{DL} approaches have increasingly been applied to jamming and interference detection, including neural network-based detection~\cite{gecgel2019jammer}, hybrid classifiers for known and unknown attacks~\cite{hong2023hybrid}, Bayesian inference-assisted classification in \gls{5G} \gls{nr}~\cite{jere2023bayesian}, and spectrogram-based supervised and unsupervised detection~\cite{varotto2024detecting}. However, many evaluations are conducted under relatively homogeneous conditions, which can overestimate deployment performance by limiting the distribution shifts encountered during evaluation. Supervised learning-based approaches also require labeled attack data before deployment, creating a barrier in environments with no prior jamming history.

Our work addresses both limitations. We benchmark under both single-environment and multi-environment \gls{loo} validation across eight heterogeneous \gls{5G} logs, directly characterizing distribution shift across deployment conditions. By evaluating both \gls{sem} and learning-based methods under identical conditions, this paper provides, to the best of our knowledge, the first comparison that isolates the contribution of the detection architecture from that of labeled data.


%% file: Methodology_V1.tex
\section{Methodology}
\label{sec:methodology}

This section describes the proposed \gls{sem}, \gls{ml} classifiers, and \gls{DL} sequence models. It also outlines the validation methodology used to assess generalization across heterogeneous deployment conditions.

\subsection{Statistical Ensemble Method (SEM)}
\gls{sem} processes per-\gls{ue} \gls{kpi} telemetry without requiring a training phase. It adapts to local signal conditions through three component detectors, which are combined using a weighted voting system~\cite{cheng2017time,ciocarlie2013detecting}. The parameters are organized into two tiers. Tier~1 parameters are based on signal-physics principles and use established default values that do not require labeled data. This represents the cold-start mode of \gls{sem}. Tier~2 denotes supervised refinement, in which selected detector, fusion, and post-processing parameters are adjusted offline through grid search once labeled data becomes available. The three detectors, categorized by tier, are detailed below. 

\textbf{Baseline Statistical Detector:} For each \gls{kpi} channel $S_k(t)$, a rolling median $\hat{\mu}_k(t)$ and rolling \gls{mad} $\hat{\sigma}_k(t)$ are computed over a centered window of $W$ samples. The modified z-score is~\cite{cheng2017time}:
\begin{equation}
z_k(t) = 0.6745 \cdot \frac{|S_k(t) - \hat{\mu}_k(t)|}{\hat{\sigma}_k(t)}.
\end{equation}
A sample is flagged if $z_k(t) > \tau_z$ for any channel $k$. The default threshold $\tau_z=3.0$ is used in the Tier~1 cold-start configuration and corresponds to the standard $3\sigma$-equivalent rule for outlier detection based on the \gls{mad}. 
This approach does not require labeled data for validation. The rolling formulation of the detector 
reduces sensitivity to long-term signal drift. When the data series is too short to allow for adaptive windowing, the method defaults to using global median and \gls{mad} statistics.

\textbf{Count-Based Detector:} The count of \gls{rsrp} samples, denoted as $C(t)$, represents the number of \gls{rsrp} samples successfully acquired during the reporting interval. Under conditions of high-gain jamming, this count can drop to zero when the receiver is unable to complete its measurements. This behavior has been consistently observed across the collected logs. The detection system flags $C(t) = 0$ without exception, and it also flags cases where the absolute difference $|C(t) - C(t-1)|$ exceeds an adaptive threshold, which is calculated based on the local rolling standard deviation. Notably, zero-count events are not present under normal operating conditions, which contributes to the high precision of this detector~\cite{ciocarlie2013detecting}. In cold-start operation, the detector uses a default minimum threshold for
the adaptive change rule, whereas this threshold can subsequently be refined during supervised grid search.

\textbf{Abrupt Change Detector:} For each channel $k$, we compute the absolute first-order difference, defined as $\Delta_k(t) = |S_k(t) - S_k(t-1)|$. Samples where $\Delta_k(t)$ exceeds the threshold $T_k = (S_k^{\max} - S_k^{\min}) \cdot p_c$ are flagged as abrupt change events. This threshold is self-normalizing to the signal's dynamic range, allowing it to adapt automatically to different \gls{kpi} magnitudes across deployment sites without requiring labeled data. The default value for $p_c$ is $0.4$, which is considered a Tier 1 parameter. Additionally, sustained measurement gaps, where no \gls{kpi} values are reported due to receiver outages, are flagged as implicit anomalies because jammer-induced receiver disruption can result in systematic measurement loss~\cite{pirayesh2022jamming}. 

\textbf{Fusion and Decision:} The three binary masks, $b_t$ 
(baseline), $c_t$ (count), and $a_t$ (abrupt), are combined through 
a weighted vote~\cite{ciocarlie2013detecting}:
\begin{equation*}
V(t) = w_{b} \cdot b_{t} + w_{c} \cdot c_{t} + w_{a} \cdot a_{t}, \quad w_{b} + w_{c} + w_{a} = 1.
\end{equation*}
Three decision masks are derived from $V(t)$. The \emph{union mask} flags any sample where at least one detector is activated: $U_t = b_t \vee c_t \vee a_t$. The \emph{weighted mask} flags samples where the confidence exceeds a threshold $\theta$: $W_t = 1$ if $V(t) \geq \theta$, and $0$ otherwise.
The \emph{hybrid mask} is 
\begin{equation*}
    H_t = c_t \vee (b_t \wedge V(t) \geq 0.5) \vee (a_t \wedge V(t) \geq 0.6),
\end{equation*} 
where $\vee$ denotes logical OR and $\wedge$ denotes logical AND. The hybrid rule gives the count detector unconditional priority, requiring the remaining detectors to reach a confidence threshold before contributing. All reported results utilize the hybrid mask, which consistently outperformed the union and weighted variants across validation conditions by reducing false alarms while preserving recall. Adjacent detections that are separated by less than a configurable gap are merged, and isolated detections that are shorter than a minimum duration are suppressed using a persistence filter. The voting weights $\{w_b, w_c, w_a\}$, the gap threshold, the persistence duration, and the threshold $\theta$ are Tier~2 parameters. These parameters have no basis in signal physics and are optimized through a grid search over the available labeled logs.

\subsection{Learning-Based Models}
We evaluate five \gls{ml} classifiers: \gls{rf}, eXtreme Gradient Boosting (XGBoost), Light Gradient Boosting Machine (LightGBM), \gls{svm}, and \gls{lr}. We also evaluate five \gls{DL} sequence models: \gls{lstm}, \gls{gru}, \gls{1dcnn}, \gls{cnn}-\gls{lstm}, and \gls{tcn}~\cite{jere2023bayesian,hong2023hybrid,varotto2024detecting}.

The learning-based methods use a common feature-engineering pipeline applied to the nine \gls{kpi} channels. The representation includes rolling statistics over windows of 5, 10, and 20 samples; lag and lag-difference features at offsets of 1, 2, 3, and 5 samples; first- and second-order changes; volatility and statistical descriptors; cross-\gls{kpi} relationships; and selected frequency-domain features. When all nine \gls{kpi} channels are available, this produces approximately $423$ features per sample. The \gls{ml} classifiers operate directly on these feature vectors, whereas the \gls{DL} models process sequences of $10$ consecutive feature vectors to capture temporal dependencies. Feature scaling is fitted using the training data and subsequently applied to validation and test data.

The \gls{ml} classifiers are trained using their respective classification objectives, with class balancing applied where supported. The \gls{DL} models use class-weighted binary cross-entropy to address label imbalance, with early stopping based on validation performance. All five \gls{ml} classifiers and all five \gls{DL} architectures are evaluated in each fold, and the model achieving the highest test-set F1-score within each family is reported as the best \gls{ml} or \gls{DL} model for that fold.

\subsection{Validation Methodology}

\emph{Single-environment validation} evaluates each log independently using a temporal split that preserves sample order. By default, $70\,\%$ of each log is used for training, $15\,\%$ for validation, and the remaining $15\,\%$ for testing. This setting represents a detector trained and evaluated under the same underlying network conditions. When necessary, the validation/test boundary is adjusted to ensure that both jamming and non-jamming samples are represented in the test partition.

\emph{Multi-environment \gls{loo} validation} holds out one complete log for testing while the remaining seven logs form the training corpus. Each log is used as the held-out test environment once. For the \gls{DL} models, a validation subset is additionally drawn from the multi-environment training corpus for early stopping. This regime represents deployment in an environment that was excluded from the training corpus and therefore provides a direct test of generalization across heterogeneous conditions.

\gls{sem} does not undergo model training in either validation regime. In cold-start mode, it operates using its default parameter configuration without labeled-data calibration. In supervised mode, its parameters are refined through grid search using available labeled data. Because \gls{sem} is not trained using the single- or multi-environment learning corpora, its operation is independent of the training regime used for the \gls{ml} and \gls{DL} models. Consequently, the cold-start \gls{sem} results are identical across the two validation settings.

%% file: Experimental_Setup_V1.tex
\section{Experimental Setup}
\label{sec:setup}
This section describes the dataset, feature representation, and evaluation metrics used to assess the proposed and learning-based detection methods under varying deployment conditions.
\subsection{Dataset}
\begin{table}[t]
    {
    \centering
    \begin{tabular}{p{0.5cm}p{0.25cm}p{1cm}p{1cm}p{1cm}p{1.25cm}p{1cm}}
    \hline
    \textbf{Log} & \textbf{\gls{ue}} & \textbf{Duration} &
    \textbf{On-time} & \textbf{Gain (dB)} & \textbf{Burst Pattern} &
    \textbf{Samples} \\
    \hline \hline
    Log1 & $3$ & $1$\,hr   & $60$\,s       & $48$ & Fixed    & $12{,}186$ \\
    Log2 & $2$ & $1$\,hr   & $60$\,s       & $48$ & Fixed    & $7{,}468$  \\
    Log3 & $2$ & $42$\,min & $60$\,s       & $48$ & Fixed    & $5{,}132$  \\
    Log4 & $1$ & $1$\,hr   & $60$\,s       & $48$ & Fixed    & $2{,}811$  \\
    Log5 & $1$ & $1$\,hr   & $30$\,s       & $48$ & Fixed    & $4{,}750$  \\
    Log6 & $1$ & $15$\,min & $60$\,s       & $45$ & Fixed    & $926$      \\
    Log7 & $1$ & $3$\,hr   & $[10,30]$\,s  & $45$ & Variable & $10{,}981$ \\
    Log8 & $1$ & $18$\,hr  & $[10,30]$\,s  & $45$ & Variable & $65{,}075$ \\
    \hline
    \end{tabular}
    \caption{Dataset summary.}
    \label{tab:dataset}}
    \vspace{-1.25em}
\end{table}

The dataset in Table~\ref{tab:dataset} includes eight experimental logs collected over approximately $25$ hours from a \gls{5G} \gls{nr} testbed operating on Band N78 under active jamming conditions~\cite{bogucka2025open}. The jamming transmission gains were set at $45$\,dB and $48$\,dB. Network traffic varied from idle to $11$ Mb/s downlink, with \gls{icmp} ping traffic included. Ground truth labels indicating the actual jamming intervals were manually verified during the experiments, allowing for parameter optimization and performance evaluation. 

The dataset encompasses three axes of variation that serve to stress-test the methods evaluated in this paper. The number of \glspl{ue}, ranging from 1 to 3, impacts signal diversity and measurement statistics, giving learning-based models a structural advantage with multi-\gls{ue} logs. The data volume, which ranges from $926$ to $65{,}075$ samples, determines the density of training examples for each deployment condition, directly testing the sample efficiency of each approach. Lastly, the burst patterns, either fixed at $30$ or $60$ seconds or variable between $10$ and $30$ seconds, introduce different levels of temporal complexity in the jamming signature. Models trained on fixed-burst logs must be able to generalize to variable-burst conditions during multi-environment validation, which represents the primary source of distribution shift in this dataset.
\subsection{\gls{kpi} Features}
Nine \gls{kpi} channels are extracted from the \gls{5G} testbed measurement reports: the minimum, first quartile (Q1), median, third quartile (Q3), maximum, and mean of \gls{rsrp}; \gls{rsrp} sample count; downlink \gls{rssi}; and uplink \gls{snr}. Together, these channels capture three principal effects of broadband jamming on the radio interface: changes in received signal strength, loss of measurement availability, and changes in the observed noise and interference level.

For the learning-based methods, the nine \gls{kpi} channels are transformed through the feature-engineering pipeline described in Section~\ref{sec:methodology}. The representation includes rolling statistics over windows of $5$, $10$, and $20$ samples; lag and lag-difference features at offsets of $1$, $2$, $3$, and $5$ samples; first- and second-order change features; statistical descriptors; cross-\gls{kpi} relationships; and selected frequency-domain features. This produces approximately $423$ features per sample when all nine \gls{kpi} channels are available. The \gls{ml} classifiers operate on individual feature vectors, whereas the \gls{DL} models use sequences of $10$ consecutive feature vectors. In contrast, \gls{sem} operates directly on the nine original \gls{kpi} channels without feature engineering.

\subsection{Evaluation Metrics}
\begin{figure}
  \centering
  \begin{subfigure}{\columnwidth}
    \includegraphics[width=\columnwidth]{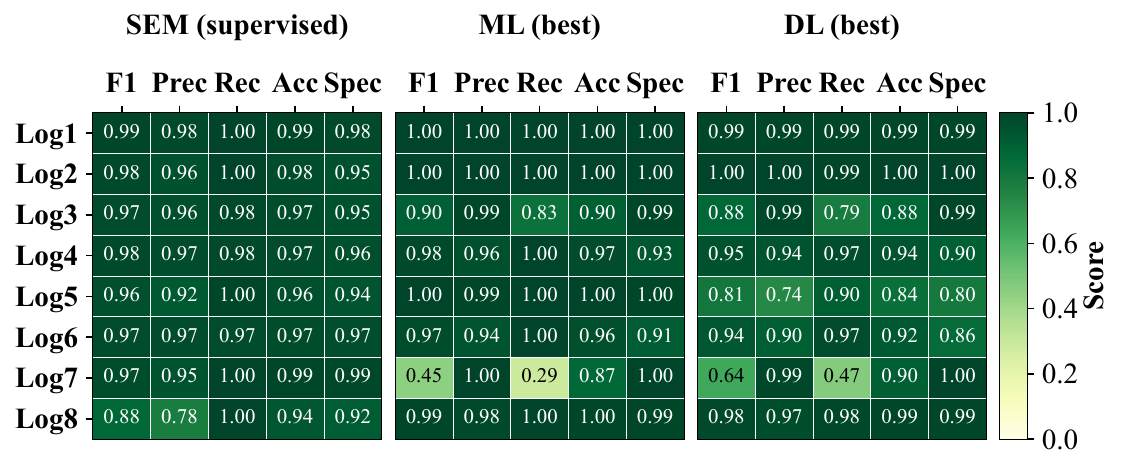}
    \caption{Single-environment validation.}
    \label{fig:sub1}
  \end{subfigure}\\
  \begin{subfigure}{\columnwidth}
    \includegraphics[width=\columnwidth]{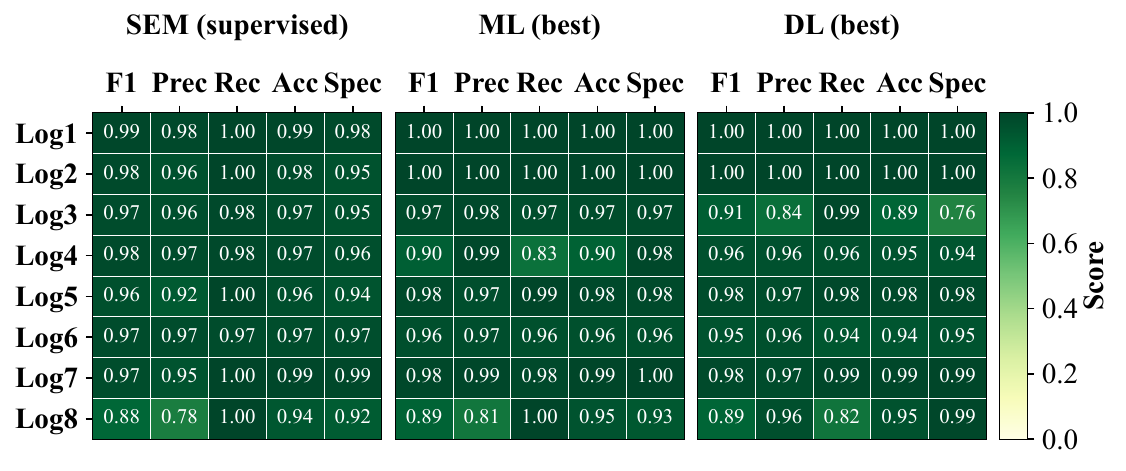}
    \caption{Multi-environment \gls{loo} validation.}
    \label{fig:sub2}
  \end{subfigure}
  \caption{Per-fold F1-score, precision, recall, accuracy, and specificity for \gls{sem} (supervised), the best-performing \gls{ml} model, and the best-performing \gls{DL} model under both validation regimes.}
  \label{fig:per_fold}
  \vspace{-1.25em}
\end{figure}
We report F1-score, precision, recall, accuracy, and specificity at the sample level. Metrics are computed separately for each held-out log to expose performance variation across deployment conditions, while mean values across the eight logs are used to summarize overall performance. F1-score captures the balance between precision and recall, whereas specificity quantifies the ability to correctly identify non-jamming samples and therefore provides an indication of false-alarm behavior. In military jamming detection, recall is particularly important because a missed jamming event can have greater operational consequences than a false alarm. We therefore interpret recall alongside F1-score throughout the results rather than relying on aggregate accuracy alone. For each fold, all five \gls{ml} classifiers and all five \gls{DL} architectures are evaluated, and the highest-F1 model within each family is reported as the best-performing \gls{ml} and \gls{DL} model, respectively.

%% file: Results_V1.tex
\section{Results}
\label{sec:results}
This section presents the overall performance, per-log distribution shift analysis, failure mode characterization, and detailed individual model breakdown.

\subsection{Overall Performance}
\begin{table*}[h]
    {
    \centering
    \begin{tabular}{llcccccl}
    \hline
    \textbf{Method} & \textbf{Regime} & \textbf{F1-score} & \textbf{Precision} 
    & \textbf{Recall} & \textbf{Accuracy} & \textbf{Specificity} 
    & \textbf{Labels?} \\
    \hline \hline
    \gls{sem} (cold-start) & Both   & $0.888$ & $0.826$ & $0.971$ 
    & $0.895$ & $0.824$ & No \\
    \gls{sem} (supervised) & Both   & $\mathbf{0.962}$ & $0.937$ & $\mathbf{0.992}$ 
    & $\mathbf{0.971}$ & $0.957$ & Yes \\
    \gls{ml} (best)        & Single & $0.911$ & $\mathbf{0.984}$ & $0.890$ & $0.963$ 
    & $\mathbf{0.978}$ & Yes \\
    \gls{ml} (best)        & Multi  & $0.962$ & $0.962$ & $0.966$ & $0.968$ 
    & $0.978$ & Yes \\
    \gls{DL} (best)        & Single & $0.897$ & $0.940$ & $0.883$ & $0.933$ 
    & $0.941$ & Yes \\
    \gls{DL} (best)        & Multi  & $0.956$ & $0.958$ & $0.958$ & $0.962$ 
    & $0.950$ & Yes \\
    \hline
    \end{tabular}
    \caption{Mean performance across 8 logs. \gls{sem} cold-start uses no labeled data; \gls{sem} supervised uses grid-search-optimized parameters. Bold entries highlight the leading value per metric column based on the unrounded mean values.}
    \label{tab:performance}}
    \vspace{-1.25em}
\end{table*}

Table~\ref{tab:performance} presents the average performance across all eight logs for \gls{sem} in cold-start mode, \gls{sem} in supervised mode, and the best results from both \gls{ml} and \gls{DL} under both validation regimes. The analysis is structured around three key findings.

\textbf{Cold-start parity with recall advantage.} \gls{sem}'s cold-start method achieves a mean F1-score of $0.888$ without labeled data, just $0.023$ points lower than single-environment \gls{ml} trained on $70\,\%$ of each log. While \gls{sem}'s cold-start precision stands at $0.826$, compared to \gls{ml}'s precision of $0.984$, the difference is largely attributed to a higher false alarm rate. This is evident from \gls{sem}'s cold-start recall of $0.971$, in contrast to $0.890$ for \gls{ml}. In military jamming detection, where missed detections are more critical than false alarms, \gls{sem}'s cold-start is the stronger baseline despite its lower F1-score.

\textbf{Grid search recovers F1-score by $+0.074$, mainly from two specific logs.} Supervised \gls{sem} achieves a mean F1-score of $0.962$, which remains consistent across both validation regimes. The largest gains over cold-start operation occur for Log~6 ($+0.195$ with $926$ samples, experiencing adaptive windowing instability at short durations) and Log~8 ($+0.151$ with an $18$-hour low-gain long duration). Log~3 improves by $+0.070$, while improvements across the remaining five logs range from approximately $+0.018$ to $+0.046$, indicating that the Tier~1 default parameters are generally robust across most deployment conditions.

\textbf{Corpus coverage drives \gls{ml} and \gls{DL} recovery.} Multi-environment training improves the \gls{ml} F1-score from $0.911$ to $0.962$ and the \gls{DL} F1-score from $0.897$ to $0.956$ compared to single-environment training. This improvement is primarily due to the diversity of the training data rather than differences in hyperparameter selection. This finding indicates that corpus coverage is a key factor limiting the performance of learning-based models when faced with distribution shifts.
\begin{figure*}[h]
    \centering
    \includegraphics[width=\textwidth]{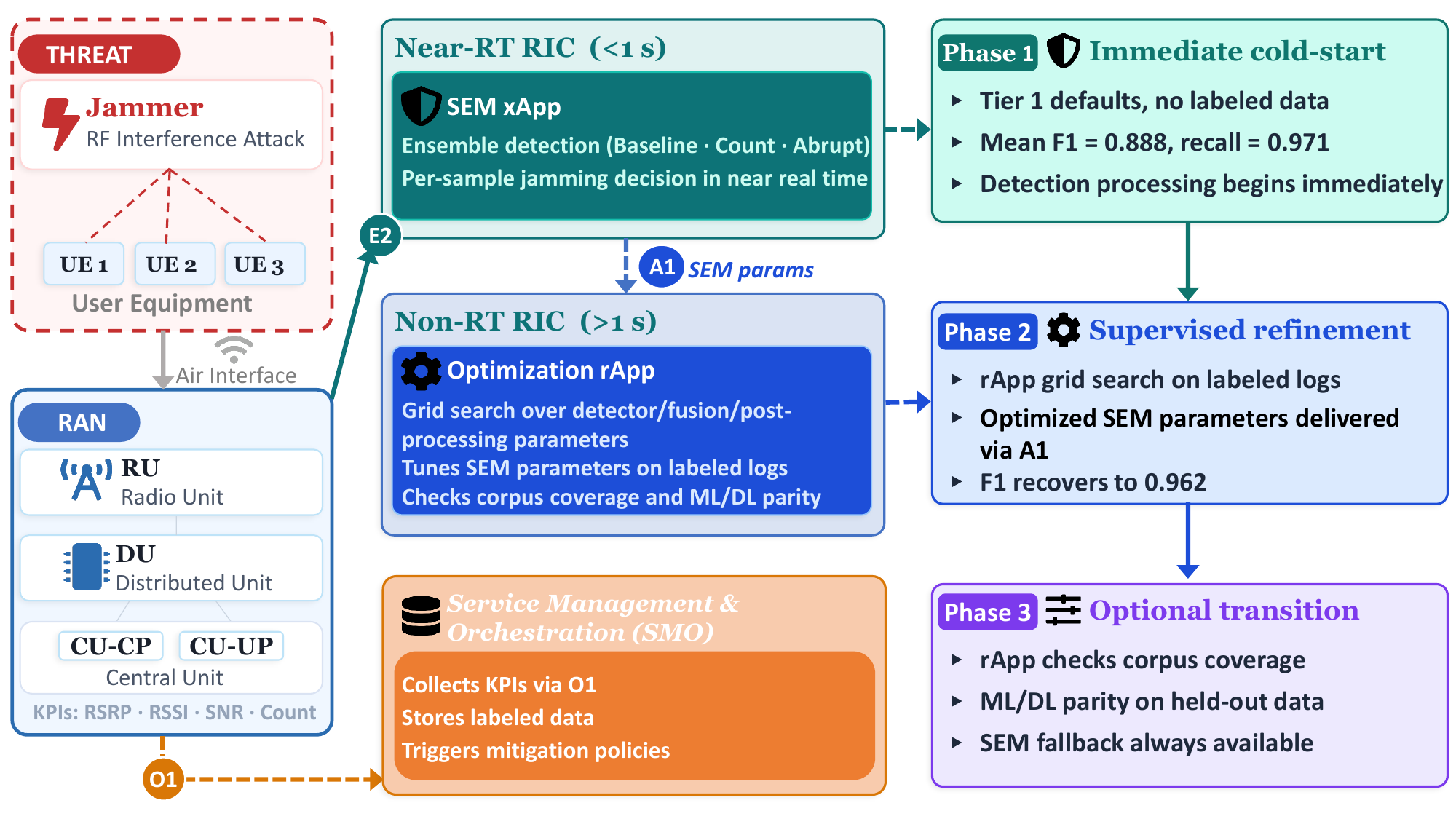}
    \caption{\gls{sem} deployment within the \gls{oran} \gls{ric} 
    architecture and three-phase deployment sequence.}
    \label{fig:deployment}
\end{figure*}
\subsection{Distribution Shift Analysis}
The overall averages in Table~\ref{tab:performance} mask significant variations across individual logs. Fig.~\ref{fig:per_fold} illustrates the per-fold performance metrics for all eight logs, revealing four distinct performance patterns.
\begin{enumerate}[leftmargin=*]
    \item \textbf{Stable high-gain conditions (Logs~$\mathbf{1}$ and $\mathbf{2}$, $\mathbf{48}$\,dB, multi-\gls{ue}, fixed $\mathbf{60}$\,s).} Supervised \gls{sem} and the learning-based methods achieve F1-scores of approximately $0.98$ or higher in both regimes. The presence of strong interference signatures, a sufficient number of samples, and regular burst patterns allows all methods to effectively detect jamming in these logs, irrespective of the training strategy used. \gls{sem}'s cold-start method achieves F1-scores between $0.93$ and $0.97$ in this context, even without any labeled data.

    \item \textbf{Sparse data and short recordings (Logs~$\mathbf{3}$, $\mathbf{4}$, and $\mathbf{6}$, $\mathbf{926}$--$\mathbf{5{,}132}$ samples, $\mathbf{15}$--$\mathbf{42}$\,min).} Supervised \gls{sem} maintains an F1-score of $0.97$-$0.98$ across all three logs. However, the cold-start performance of \gls{sem} degrades significantly on Log~6, with an F1-score of $0.78$ and precision of $0.69$. In this case, the rolling window of $W=50$ samples accounts for approximately $5\,\%$ of the $926$-sample recording, making the adaptive baseline more sensitive to the short recording duration. In contrast, both \gls{ml} and \gls{DL} approaches experience moderate degradation in the single-environment scenario for Log~3 (\gls{ml}: $0.90$, \gls{DL}: $0.88$). Under multi-environment training, \gls{ml} recovers to $0.97$, while \gls{DL} reaches $0.91$. Recovery is also not complete for Log~4, where the multi-environment \gls{ml} performance is $0.90$, indicating that the operating conditions of the held-out log are not fully represented by the remaining training corpus.

    \item \textbf{Variable burst pattern, distribution shift regime (Log~$\mathbf{7}$, $\mathbf{[10,30]}$\,s bursts, $\mathbf{45}$\,dB).} Log~7 generates the most significant operational results. In a single-environment training scenario, \gls{ml} performance degrades to an F1-score of $0.45$ with a recall of $0.29$, failing to detect $71\,\%$ of the labeled jamming samples. Meanwhile, \gls{DL} achieves an F1-score of $0.64$ with a recall of $0.47$. In contrast, \gls{sem} in cold-start mode achieves an F1-score of $0.93$ with a recall of $0.96$ across both regimes; this represents a $0.48$-point advantage in F1-score and a $0.67$-point advantage in recall over single-environment \gls{ml} for the most challenging distribution-shift case. Under multi-environment training, both learning-based methods recover to an F1-score of approximately $0.98$, comparable to the $0.97$ F1-score of supervised \gls{sem}. This recovery underscores that diverse training data can address distribution shifts when available, but prior exposure to similar variable-burst behavior is necessary before deployment.

    \item \textbf{Large dataset with underrepresented conditions (Log~$\mathbf{8}$, $\mathbf{18}$\,hr, $\mathbf{[10,30]}$\,s bursts, $\mathbf{45}$\,dB, $\mathbf{65{,}075}$ samples).} Log~8 demonstrates that a larger training corpus does not necessarily improve generalization when the target operating condition is insufficiently represented. The F1-score for \gls{ml} declines from $0.99$ in single-environment validation to $0.89$ in multi-environment validation. Similarly, \gls{DL} declines from $0.98$ to $0.89$, while supervised \gls{sem} remains at approximately $0.88$ across both regimes. The training corpus for the multi-environment fold of Log~8 consists of seven shorter logs that do not adequately represent its $18$-hour low-gain jammer structure with variable bursts. This is a dataset coverage limitation for the learning-based methods, indicating that multi-environment training is not universally advantageous. Its effectiveness depends on whether the training data adequately covers the conditions present in the target deployment environment.
\end{enumerate}

Taken together, Logs~7 and 8 illustrate two complementary aspects of cross-environment generalization. Log~7 represents a recoverable distribution shift: learning-based methods perform poorly under single-environment training but recover when the training corpus provides broader exposure to heterogeneous operating conditions. Log~8 presents the opposite case. Although the multi-environment training corpus contains substantially more aggregate samples, those samples do not sufficiently represent the long-duration, low-gain, variable-burst behavior of the held-out environment. Thus, the two logs demonstrate that training-set size and training-set coverage are distinct quantities. Increasing the amount of labeled data is beneficial only when the resulting corpus captures the conditions likely to occur after deployment.

\subsection{Failure Mode Analysis}
Table~\ref{tab:log7} highlights the key differences between \gls{sem} and learning-based methods in the context of Log~7, which presents the most challenging distribution shift. \gls{sem}'s cold-start mode demonstrates a safe failure mode: while precision decreases, recall remains intact, which means that jamming samples are still detected but with some false alarms. In contrast, single-environment \gls{ml} and \gls{DL} fail in an unsafe manner; recall drops to $0.29$ and $0.47$, respectively, resulting in a significant number of undetected jamming samples. In military \gls{c2} networks, failing to detect jamming activity is an operationally unacceptable outcome.

\begin{table}[t]
    {
    \centering
    \begin{tabular}{llccc}
    \hline
    \textbf{Method} & \textbf{Regime} & \textbf{F1} & \textbf{Precision} 
    & \textbf{Recall} \\
    \hline \hline
    \gls{sem} cold-start & Both   & $0.93$ & $0.90$ & $0.96$ \\
    \gls{sem} supervised & Both   & $0.97$ & $0.95$ & $1.00$ \\
    \gls{ml} (best)      & Single & $0.45$ & $1.00$ & $0.29$ \\
    \gls{ml} (best)      & Multi  & $0.98$ & $0.99$ & $0.98$ \\
    \gls{DL} (best)      & Single & $0.64$ & $0.99$ & $0.47$ \\
    \gls{DL} (best)      & Multi  & $0.98$ & $0.97$ & $0.99$ \\
    \hline
    \end{tabular}
    \caption{Per-method performance on Log~7.}
    \label{tab:log7}}
    \vspace{-1.25em}
\end{table}

\subsection{Individual Model Analysis}
Beyond the best-per-family results, the individual model analysis reveals substantial variation within both learning-based families. In single-environment training, LightGBM achieves the highest F1-score of $0.91$ among \gls{ml} classifiers, while the \gls{1dcnn} scores $0.88$ among \gls{DL} models. Multi-environment training substantially boosts performance, with \gls{rf} reaching an F1-score of $0.96$ and XGBoost reaching $0.94$. \gls{cnn}-based models, including \gls{1dcnn} and \gls{cnn}-\gls{lstm}, also score $0.94$. However, two models are notable outliers: \gls{svm} scores $0.20$, while \gls{tcn} scores only $0.10$ under multi-environment training.

%% file: System_Integration_V1.tex
\section{O-RAN Deployment Implications}
\label{sec:integration}
Fig.~\ref{fig:deployment} illustrates \gls{sem}'s deployment within the \gls{oran} \gls{ric} architecture. \gls{sem}'s ensemble detector runs as an \gls{xApp} on the near-\gls{rt} \gls{ric}, receiving live \gls{kpi} telemetry  via the E2 interface and producing per-sample jamming decisions in near real time. The Non-\gls{rt} \gls{ric} hosts the optimization \gls{rApp}, which runs grid search over selected detector, fusion, and post-processing parameters using available labeled logs and delivers the optimized parameters to the 
\gls{xApp} through the A1 interface~\cite{kryszkiewicz2023open,bogucka2025open}. This effectively separates real-time detection from offline optimization within the \gls{oran} stack.

\textbf{Phase 1: Immediate Cold-Start:} \gls{sem} is deployed with Tier~1 defaults upon initial activation, allowing detection processing to begin immediately. Although precision is lower in this mode (mean $0.826$), there are elevated false alarm rates in very short recordings and during long-duration low-gain conditions. However, recall remains $\geq 0.88$ across all conditions. 

\textbf{Phase 2: Supervised Refinement:} As labeled incidents accumulate, the \gls{rApp} performs a grid search and delivers optimized parameters through A1 without interrupting live detection. The improvements are largest for Log~6-type short recordings, which show a gain of $+0.195$ in F1-score, and for Log~8-type long-duration low-gain conditions, which show a gain of $+0.151$ in F1-score. Log~3 improves by $+0.070$, while gains across the remaining five logs range from approximately $+0.018$ to $+0.046$.

\textbf{Phase 3: Optional Transition:} Once the dataset sufficiently covers the target deployment conditions, the \gls{rApp} assesses whether a learning-based alternative has achieved performance parity with \gls{sem}, as evaluated on held-out data. Multi-environment \gls{ml} achieves a mean F1-score of $0.962$, essentially matching the supervised \gls{sem} mean F1-score of $0.962$. \gls{sem} remains available as a fallback option, given that multi-environment training may not always be beneficial: Log~8 illustrates 
that when the training dataset does not adequately cover the target deployment conditions, performance can decline regardless of dataset size.

%% file: Conclusion_V1.tex
\section{Conclusion}
\label{sec:conclusion}

This paper presents \gls{sem}, a method for detecting jamming in
\gls{oran}-native environments. \gls{sem} is evaluated against five
\gls{ml} classifiers and five \gls{DL} sequence models using eight
\gls{5G} logs. The main finding is that supervised \gls{sem} outperforms
single-environment learning-based methods in mean F1-score, while
cold-start \gls{sem} provides high recall without requiring labeled data.
While learning-based models struggle with recall during distribution shifts,
\gls{sem} tends to preserve detection capability at the cost of additional
false alarms. Multi-environment \gls{ml} achieves a mean F1-score of
$0.962$, essentially matching supervised \gls{sem} at $0.962$, although
this similarity in mean performance masks substantial variation across
individual logs. Additionally, performance improvements under
multi-environment training from $0.911$ to $0.962$ for \gls{ml} and from
$0.897$ to $0.956$ for \gls{DL} demonstrate the importance of diverse
training data, indicating that corpus coverage is a key limitation for
learning-based methods in heterogeneous environments.

%% file: Acknowledgement.tex
\section*{Acknowledgment}
This work is partially funded by the Department of Homeland Security’s Science and Technology Directorate (S\&T). U.S. Government work, not subject to U.S. Copyright.